\documentclass{article}
\usepackage{spconf,amsmath,graphicx,hyperref}
\usepackage{amssymb}  
\usepackage{xcolor}
\usepackage{multirow} 
\usepackage{etoolbox} 
\usepackage{overpic} 
\apptocmd{\thebibliography}{%
  \setlength{\itemsep}{0pt}%
  \setlength{\parskip}{0pt}%
  \setlength{\parsep}{0pt}%
}{}{}

\usepackage{booktabs}
\usepackage{siunitx}
 
\usepackage{enumitem}
\usepackage[table]{xcolor}
\usepackage{printlen}
\usepackage{layouts}
\usepackage{caption}
\title{Improving Multimodal Brain Encoding Model with Dynamic Subject-awareness Routing}
\name{Xuanhua Yin \qquad
      Runkai Zhao$^{\star}$ \thanks{* Corresponding authors} \qquad
      Weidong Cai $^{\star}$\footnotemark[1]}
       
\address{School of Computer Science, The University of Sydney, Australia \\
{\tt \{xuanhua.yin,runkai.zhao,tom.cai\}@sydney.edu.au} }
\begin{document}
%
\maketitle
\begin{abstract}
Naturalistic fMRI encoding must handle multimodal inputs, shifting fusion styles, and pronounced inter-subject variability. We introduce AFIRE (Agnostic Framework for Multimodal fMRI Response Encoding), an agnostic interface that standardizes time-aligned post-fusion tokens from varied encoders, and MIND, a plug-and-play Mixture-of-Experts decoder with a subject-aware dynamic gating. Trained end-to-end for whole-brain prediction, AFIRE decouples the decoder from upstream fusion, while MIND combines token-dependent Top-K sparse routing with a subject prior to personalize expert usage without sacrificing generality. Experiments across multiple multimodal backbones and subjects show consistent improvements over strong baselines, enhanced cross-subject generalization, and interpretable expert patterns that correlate with content type. The framework offers a simple attachment point for new encoders and datasets, enabling robust, plug-and-improve performance for naturalistic neuroimaging studies. Code is available at \url{https://github.com/xuanhuayin/MIND}.
\end{abstract}
\vspace{-1em}
\section{Introduction}
\vspace{-0.5em}
\label{sec:intro}
Real-world perception is intrinsically multisensory: auditory, visual, and linguistic cues co-occur rather than arriving in isolation. Converging evidence shows that these cues are integrated across distributed cortex, not only in heteromodal association areas but also via modulatory influences within so-called ``unimodal’’ cortices. This pattern is consistent with population codes pooling information across modalities \cite{Stein2008}. This view motivates encoding models that operate directly on multimodal inputs, targeting unified neural prediction rather than stitching together separate unimodal pipelines \cite{bao2025mindsimulator}.

\begin{figure}[t]
\centering
\includegraphics[width=\linewidth]{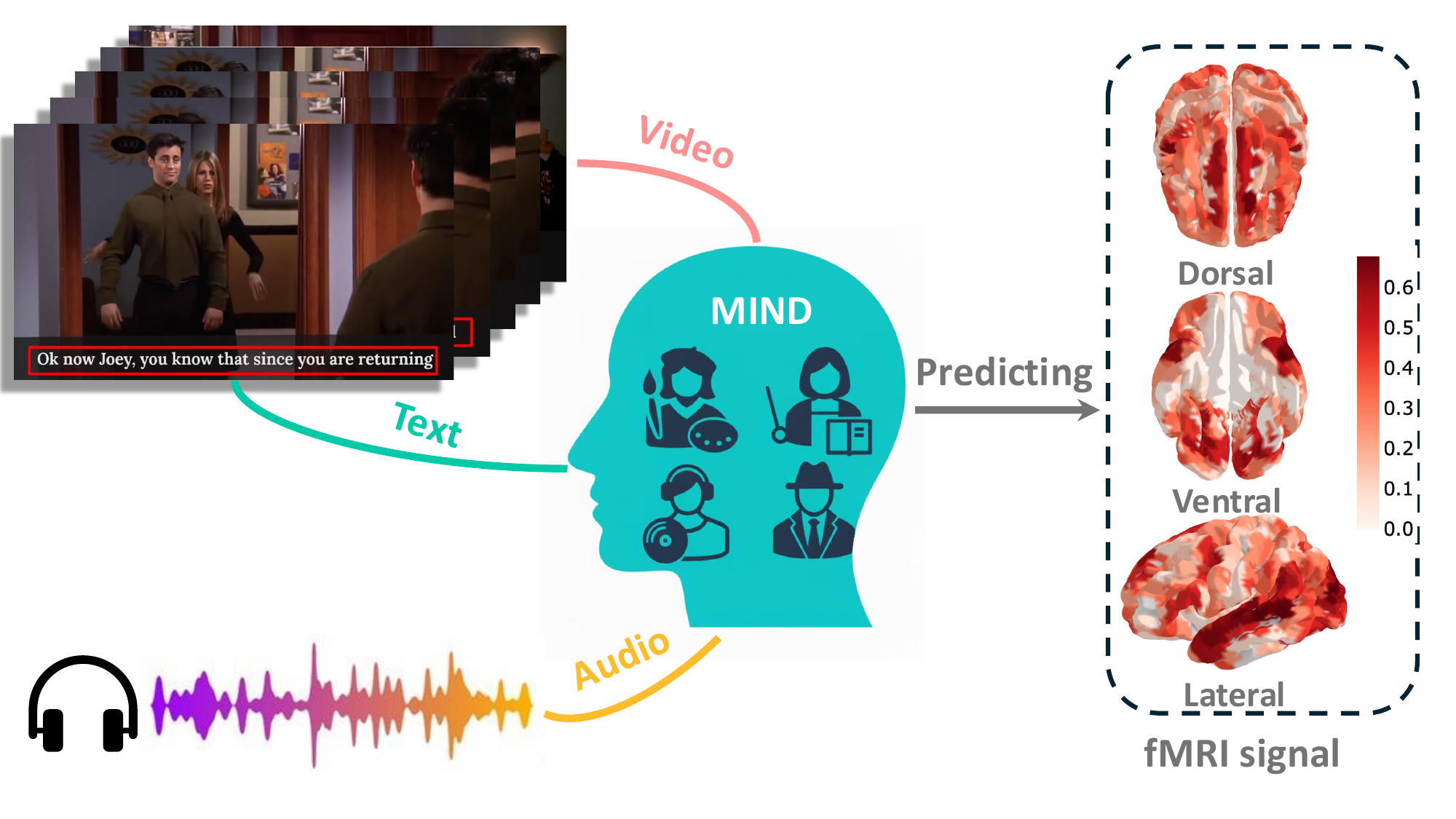}
\caption{\textbf{AFIRE pipeline for fMRI prediction.} The brain integrates multimodal information concurrently over time. AFIRE mirrors this process by exposing time-aligned, fusion-agnostic tokens to a subject-aware dynamic decoder (MIND), enabling whole-brain fMRI prediction without backbone-specific tailoring.}
\vspace{-1mm} 
\label{fig:intro}
\end{figure}
A central challenge in naturalistic multimodal brain encoding is that identical audiovisual and linguistic stimuli can evoke systematically different fMRI responses across individuals. Such inter-subject variability is not merely noise; it reflects subject-dependent attention, interpretation, and cross-modal integration. As a result, a single population-level decoder trained to predict an average response can blur subject-specific structure and generalize poorly to unseen participants.

In single-modality settings, cross-subject variability is often mitigated by aligning subjects into a shared representational space (e.g., hyperalignment \cite{Haxby2011,Guntupalli2016}) or by modeling a shared latent response as in the Shared Response Model (SRM) \cite{Chen2015}, which helps disentangle common structure from idiosyncratic topographies and improves transfer.
Under naturalistic multimodal stimulation, however, variability is not only a static alignment issue but also a dynamic fusion issue, since individuals can differ in how they weight, bind, and gate multimodal cues over time.
Standard alignments therefore fail to capture subject-specific fusion patterns, making alignment alone insufficient for robust cross-subject prediction.
Recent community efforts underscore this gap and call for multimodal-generalizable brain-encoding models \cite{al,qiu2025mindllm,dai2025mindaligner}.
This motivates architectures that impose controlled subject-specific structure on top of a shared stimulus representation, separating common structure from individual differences while remaining fusion-agnostic, interpretable, and scalable.

To address these challenges (Fig.~\ref{fig:intro}), we propose the \textbf{A}gnostic \textbf{F}ramework for Multimodal fMR\textbf{I} \textbf{R}esponse \textbf{E}ncoding (\textbf{AFIRE}), a plug-in interface that takes fused/joint multimodal features from diverse backbones and projects them into a shared, time-aligned token space for downstream decoding.
By decoupling upstream fusion from downstream prediction, AFIRE enables a stable, subject-aware decoder and improves cross-backbone transfer and inter-subject generalization.
Within AFIRE, we introduce the \textbf{M}ixture-of-Experts \textbf{In}tegrated \textbf{D}ecoder (\textbf{MIND}), a sparse MoE that maps tokens to whole-brain responses, where each subject is parameterized by a learnable mixture over experts.
We further propose \textbf{S}ubject-\textbf{a}ware \textbf{D}ynamic \textbf{Gating} (\textbf{SADGate}), which combines token-dependent scores with a subject prior and applies sparse Top-$K$ selection to produce subject- and token-conditioned expert weights.
We evaluate on \textit{Algonauts 2025} \cite{al} using three feature-fusion backbones—TRIBE \cite{d2025tribe}, ImageBind \cite{girdhar2023imagebind}, and Qwen2.5-Omni \cite{xu2025qwen2}—and report parcel-level Pearson $r$, Spearman $\rho$, $R^2$, and \textit{Inter-Subject Generalization} (ISG).
Our contributions are summarized as follows:
\begin{itemize}[leftmargin=*,
                itemsep=0.25em,  
                topsep=0.3em,    
                parsep=0pt,      
                partopsep=0pt]
\item We propose the Agnostic Framework for Multimodal fMRI Response Encoding, \textit{\textbf{AFIRE}}, an agnostic, plug-and-play framework that exposes a uniform post-fusion interface and enables whole-brain encoding across diversity fusion backbones.
\item We design the Mixture-of-Experts Integrated Decoder, \textit{\textbf{MIND}}, operating on AFIRE’s post-fusion features, and incorporate a subject dynamic routing module, \textit{\textbf{SADGate}}, combining token-dependent scores with a subject prior router and sparse Top-$K$ selection. Together with specialized expert heads and weighted late fusion, these components improve both generalization and personalization.
\item We conduct extensive experiments on the \textit{Algonauts 2025} benchmark with the Trimodal Brain Encoder (TRIBE), ImageBind and Qwen2.5-Omni encoders, achieving mean gains of \textbf{+0.067} ($r$), \textbf{+0.065} ($\rho$), \textbf{+0.028} ($R^2$), and \textbf{+0.063} (\textit{Inter-Subject Generalization}, \textbf{ISG}) over each baseline.
\end{itemize}

\vspace{-1.2em}
\section{METHODOLOGY}
\vspace{-0.7em}
\label{sec:method}

\begin{figure*}[t]
\centering
\includegraphics[width=\linewidth]{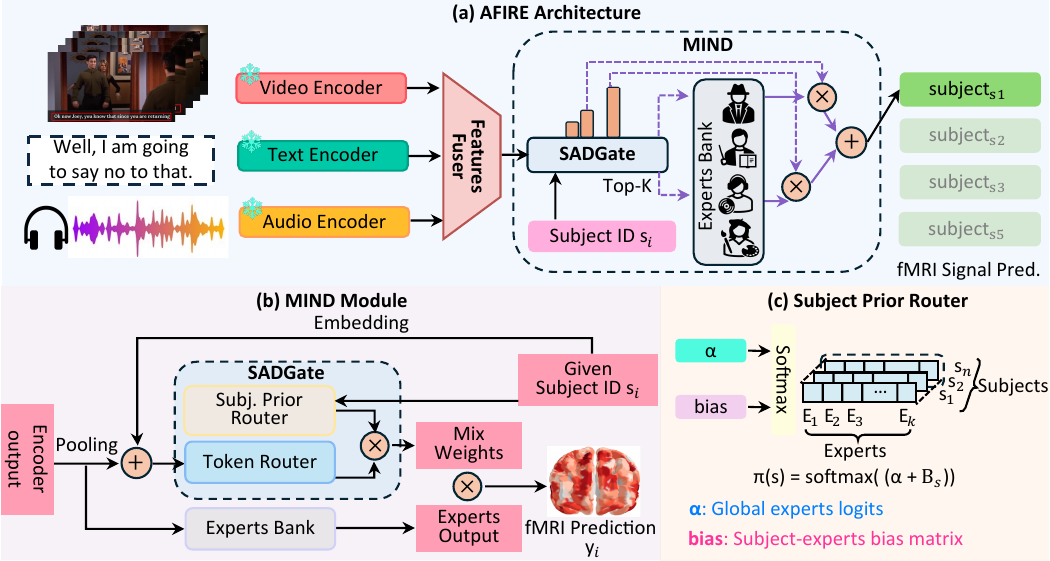}
\caption{\textbf{Overall pipeline.}
\textbf{(a) AFIRE:} a fusion-agnostic post-fusion interface that projects TR-aligned features into a shared token space, decoupling upstream fusion from decoding.
\textbf{(b) MIND:} a subject-aware sparse MoE decoder operating on shared tokens for robust backbone transfer.
\textbf{(c) Subject prior router:} a global logit vector $\alpha$ and subject--expert bias $B$ define $\pi(s)$.
Top-$K$ selection and normalization yield sparse weights $\hat{w}_t$ for stable, subject-adaptive routing.}
\vspace{-1.5mm}
\label{fig:pipeline}
\end{figure*}

We adopt \textbf{AFIRE} as an agnostic interface that standardizes heterogeneous multimodal backbones into TR-synchronous post-fusion tokens and supports end-to-end training \cite{al,tang2023transfer,oota2025multi,xia2024umbrae}.
On this interface, we deploy \textbf{MIND}, a sparse Mixture-of-Experts decoder for whole-brain fMRI, together with its subject-aware gating module \textbf{SADGate} to model inter-subject variability \cite{Nastase2020}.
By separating upstream fusion from downstream decoding, AFIRE allows MIND to impose stable subject-aware inductive biases while remaining compatible with image-aligned, Transformer-fused, and LLM-centric encoders \cite{girdhar2023imagebind,d2025tribe,xu2025qwen2,ChenDLWH25}.

\vspace{-0.7em}
\subsection{Data Temporal Alignment}
\vspace{-0.3em}
\label{sec:align}
Upstream multimodal features are sampled at 2\,Hz, while fMRI is collected at TR resolution.
We aggregate 2\,Hz frames within each TR bin to obtain TR-synchronous post-fusion tokens $\{z_t\}_{t=1}^{T}$ with $z_t\in\mathbb{R}^{D}$,
which serve as AFIRE inputs to the decoder,
where $t$ indexes TRs, $T$ is the sequence length in TR steps, and $D$ is the AFIRE token dimensionality.

\vspace{-0.5em}
\subsection{AFIRE: An Agnostic Framework}
\vspace{-0.3em}
\label{sec:afire}

\noindent\textbf{Pipeline \& Interface.}
As shown in Fig.~\ref{fig:pipeline}a, AFIRE takes TR-aligned, per-layer features from heterogeneous fusion backbones and projects them into a shared token space via a lightweight projector.
A fusion operator then merges modalities into a single token stream to form a standardized post-fusion sequence $\{z_t\}$.
We further apply a temporal MLP with positional encoding and layer normalization to capture short-range dependencies while keeping the interface backbone-agnostic.
This interface enables a unified decoder across image-aligned, Transformer-fused, and LLM-centric backbones~\cite{girdhar2023imagebind,d2025tribe,xu2025qwen2}.

\noindent\textbf{Training Scheme.}
AFIRE is trained end-to-end with the decoder under the same reconstruction objective.
Gradients flow through the decoder, experts, and routers into the AFIRE projector and temporal module.
This joint training stabilizes $\{z_t\}$ and supports transfer across upstream fusion designs~\cite{al,tang2023transfer}.

\vspace{-0.5em}
\subsection{SADGate: Subject-aware Dynamic Gating}
\vspace{-0.3em}
\label{sec:SADGate}
SADGate combines a Token Router and a Subject Prior Router to produce a subject- and token-conditioned Top-$K$ expert mixture,
where $s\in\{1,\dots,S\}$ indexes subjects, $S$ is the number of subjects, $E$ is the number of experts, and $K$ is the number of active experts per token.

\noindent\textbf{Token Router.}
For subject $s$ and token $z_t\in\mathbb{R}^{D}$ at TR $t$, we use a learnable subject embedding $e_{\mathrm{subj}}(s)$ to compute token-dependent routing weights
\begin{equation}
\tilde{z}_t = z_t + e_{\mathrm{subj}}(s), \quad
g_t = W_r \tilde{z}_t + b_r, \quad
p_t = \mathrm{softmax}(g_t).
\end{equation}
where $e_{\mathrm{subj}}(s)\in\mathbb{R}^{D}$, $W_r\in\mathbb{R}^{E\times D}$, $b_r\in\mathbb{R}^{E}$, and $g_t,p_t\in\mathbb{R}^{E}$.
This path adapts expert usage to transient stimulus context and local temporal cues~\cite{Nastase2020}.

\noindent\textbf{Subject Prior Router.}
To capture persistent expert preferences and stabilize allocation (Fig.~\ref{fig:pipeline}c), we maintain a global expert logit vector $\alpha\in\mathbb{R}^{E}$ and a subject--expert bias matrix $B\in\mathbb{R}^{S\times E}$
\begin{equation}
\pi(s) = \mathrm{softmax}\!\big(\alpha + B_{s,:}\big)\in\Delta^{E-1},
\end{equation}
where $B_{s,:}\in\mathbb{R}^{E}$ is the $s$-th row and $\Delta^{E-1}$ is the $(E\!-\!1)$-simplex.
We combine both routers and apply sparse Top-$K$ selection
\begin{equation}
u_t = p_t \odot \pi(s), \quad
\hat{w}_t = \mathrm{Normalize}\!\big(\mathrm{Top}\text{-}K(u_t)\big)\in\mathbb{R}^{E},
\end{equation}
where $\odot$ denotes element-wise multiplication, $\mathrm{Top}\text{-}K(\cdot)$ keeps the $K$ largest entries, and $\mathrm{Normalize}(\cdot)$ renormalizes the remaining entries to sum to one.
We denote the $e$-th component by $\hat{w}_{t,e}$.
By factoring routing into a stable subject prior and a token-adaptive gate, SADGate encourages experts to capture reusable patterns while calibrating subject-level gain.
This design follows sparse MoE practice~\cite{Shazeer2017,Fedus2022,lepikhin2021gshard} and supports subject-specific adaptation~\cite{Finn2015}.

\vspace{-0.5em}
\subsection{MIND: Mixture-of-Experts Integrated Decoder}
\vspace{-0.3em}
\label{sec:mind}
MIND instantiates $E$ MLP experts $f_e:\mathbb{R}^{D}\!\to\!\mathbb{R}^{O}$ for whole-brain prediction~\cite{Schaefer2018},
where $O$ is the number of output parcels or ROIs (e.g., $O{=}1000$ Schaefer parcels).
Given token $z_t$, the prediction is
\begin{equation}
y_t = \sum_{e=1}^{E} \hat{w}_{t,e}\, f_e(z_t),
\qquad \hat{w}_{t,e}\ge 0, \quad \sum_{e=1}^{E}\hat{w}_{t,e}=1.
\label{eq:mind}
\end{equation}
where $y_t\in\mathbb{R}^{O}$ is the predicted parcel-wise response at TR $t$.
Top-$K$ sparse routing yields interpretable subject-specific expert mixtures while keeping expert heads modular.

\vspace{-0.5em}
\subsection{Learning Objective and Regularization}
\vspace{-0.3em}
\label{sec:loss}
We optimize AFIRE, routers, subject embeddings and priors, and expert heads end-to-end using
\begin{equation}
\mathcal{L}
=
\underbrace{\mathcal{L}_{\mathrm{rec}}}_{\text{MSE}}
+
\beta\,\underbrace{\mathcal{R}_{\mathrm{lb}}}_{\text{load balancing}}
+
\lambda\,\|B\|_2^2 .
\end{equation}
where $\beta$ and $\lambda$ are scalar weights and $B$ is the subject--expert bias in Eq.~(2).
We define
\begin{equation}
\mathcal{L}_{\mathrm{rec}}=\frac{1}{T}\sum_{t=1}^{T}\|y_t-y_t^{\ast}\|_2^2,
\end{equation}
where $y_t^{\ast}\in\mathbb{R}^{O}$ is the ground-truth parcel-wise fMRI response at TR $t$.
$\mathcal{R}_{\mathrm{lb}}$ reduces expert collapse and routing volatility in sparse MoE~\cite{Fedus2022,Shazeer2017,lepikhin2021gshard}.
We measure expert importance by $\mathrm{Imp}_e=\sum_{t} u_{t,e}$ and expert load by $\mathrm{Load}_e=\sum_{t}\mathbb{I}[\hat{w}_{t,e}>0]$.
The $\ell_2$ penalty on $B$ constrains subject-specific drift while preserving personalization.

\vspace{-1em}
\section{Experiments and Results}

\vspace{-1em}
\subsection{Experimental Setup}
\vspace{-0.3em}

\noindent\textbf{Datasets and Optimization.}
We use \textit{Algonauts 2025}~\cite{al} with 2\,Hz video/audio/text features aligned to fMRI TRs.
We evaluate four subjects ($S_{1}$, $S_{2}$, $S_{3}$, $S_{5}$) and parcellate cortex into $O{=}1000$ regions using the Schaefer atlas~\cite{Schaefer2018}.
All models share the same per-subject, per-episode stratified 90/10 train/val split.
We train with AdamW and a OneCycle scheduler.
Peak learning rate and weight decay are selected by validation grid search.
Each sample is a 100-TR window with a 50-TR stride, and 2\,Hz features are averaged within each TR bin.

\noindent\textbf{Backbones and Plug-and-improve Protocol.}
We evaluate three fusion backbones, TRIBE, ImageBind, and Qwen2.5-Omni~\cite{WangSAFGWYW24}.
To test plug-and-improve, we keep the window length, hidden size, and training protocol fixed across methods.
We only swap the upstream encoder latents and attach the same MIND decoder, isolating decoder effects under each fusion style.

\noindent\textbf{Evaluation Metrics.}
We report Pearson \(r\), Spearman \(\rho\), \(R^2\), and inter-subject generalization (ISG).
Scores are computed parcel-wise on the validation set and averaged over parcels, episodes, and subjects.
ISG trains with leave-one-subject-out and reports \(r\) on the held-out subject.

\begin{figure}[t]
\centering
\includegraphics[width=0.86\linewidth,keepaspectratio]{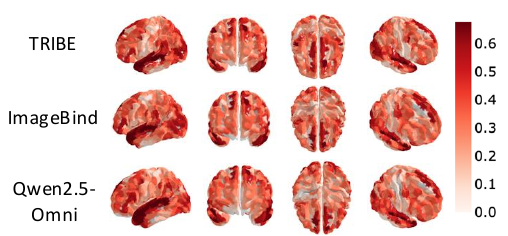}
\caption{\textbf{Parcel-wise prediction–measurement correlation on \textit{Friends} S6E5.}
TRIBE, ImageBind, and Qwen2.5-Omni (rows; all with \textit{MIND}) show similar spatial patterns of Pearson \textbf{r} between predicted and measured fMRI, indicating fusion-agnostic robustness.}
\label{fig:vis1}
\vspace{-2pt}
\end{figure}

\vspace{-0.5em}
\subsection{Main Results}
\vspace{-0.3em}

\begin{table}[t]
\centering
\setlength{\tabcolsep}{4pt}
\renewcommand{\arraystretch}{0.92}
\setlength{\aboverulesep}{0.35ex}
\setlength{\belowrulesep}{0.35ex}
\small
\caption{Mean validation performance (All Episodes; $S_{1}$--$S_{5}$) for TRIBE, ImageBind, and Qwen2.5-Omni with MLP/MMoE/MIND decoders; $\Delta$ is relative to each baseline.}
\label{tab:improvements}
\begin{tabular}{@{}ccccc@{}}
\toprule
\textbf{Methods} & \textbf{r} & \textbf{\boldmath$\rho$\unboldmath} & \textbf{$\mathbf{R}^2$} & \textbf{ISG} \\
\midrule
\multicolumn{5}{c}{\textit{TRIBE}~\cite{d2025tribe}} \\
\midrule
TRIBE (Baseline)      & 0.256 & 0.240 & 0.081 & 0.187 \\
w. MLP Decoder      & 0.247 & 0.228 & 0.069 & 0.189 \\
w. MMoE Decoder     & 0.267 & 0.252 & 0.087 & 0.198 \\
\multicolumn{1}{@{}c}{\cellcolor{gray!15}\textbf{w. MIND}} &
\multicolumn{1}{c}{\cellcolor{gray!15}\textbf{0.273}} &
\multicolumn{1}{c}{\cellcolor{gray!15}\textbf{0.259}} &
\multicolumn{1}{c}{\cellcolor{gray!15}\textbf{0.092}} &
\multicolumn{1}{c@{}}{\cellcolor{gray!15}\textbf{0.241}} \\
$\Delta$ (vs. Baseline) & {+0.017} & {+0.019} & {+0.011} & {+0.054} \\
\midrule
\multicolumn{5}{c}{\textit{ImageBind}~\cite{girdhar2023imagebind}} \\
\midrule
ImageBind (Baseline)  & 0.131 & 0.121 & 0.026 & 0.097 \\
w. MLP Decoder      & 0.139 & 0.120 & 0.027 & 0.139 \\
w. MMoE Decoder     & 0.198 & 0.181 & 0.052 & 0.147 \\
\multicolumn{1}{@{}c}{\cellcolor{gray!15}\textbf{w. MIND}} &
\multicolumn{1}{c}{\cellcolor{gray!15}\textbf{0.221}} &
\multicolumn{1}{c}{\cellcolor{gray!15}\textbf{0.203}} &
\multicolumn{1}{c}{\cellcolor{gray!15}\textbf{0.064}} &
\multicolumn{1}{c@{}}{\cellcolor{gray!15}\textbf{0.162}} \\
$\Delta$ (vs. Baseline) & {+0.090} & {+0.082} & {+0.038} & {+0.065} \\
\midrule
\multicolumn{5}{c}{\textit{Qwen2.5-Omni}~\cite{xu2025qwen2}} \\
\midrule
Qwen2.5-Omni (Baseline) & 0.125 & 0.130 & 0.025 & 0.103 \\
w. MLP Decoder         & 0.140 & 0.132 & 0.031 & 0.144 \\
w. MMoE Decoder        & 0.201 & 0.183 & 0.049 & 0.144 \\
\multicolumn{1}{@{}c}{\cellcolor{gray!15}\textbf{w. MIND}} &
\multicolumn{1}{c}{\cellcolor{gray!15}\textbf{0.220}} &
\multicolumn{1}{c}{\cellcolor{gray!15}\textbf{0.205}} &
\multicolumn{1}{c}{\cellcolor{gray!15}\textbf{0.059}} &
\multicolumn{1}{c@{}}{\cellcolor{gray!15}\textbf{0.162}} \\
$\Delta$ (vs. Baseline)    & {+0.095} & {+0.075} & {+0.034} & {+0.059} \\
\bottomrule
\end{tabular}
\normalsize
\end{table}

\begin{figure}[t]
\centering
\includegraphics[width=\linewidth]{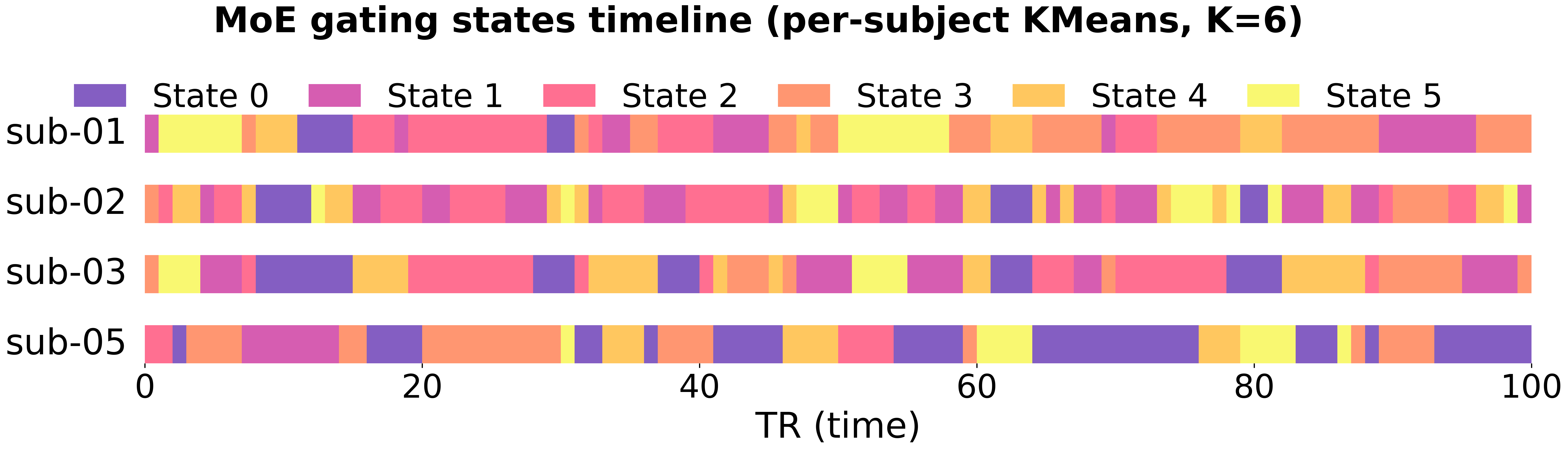}
\caption{\textbf{Subject routing dynamics (first 100 TRs).} Same-episode expert weights for $S_{1}$, $S_{2}$, $S_{3}$, $S_{5}$ (colors denote experts). Weight curves over time indicate subject-specific preferences modulated by token-dependent signals, showing that \textit{MIND} captures inter-subject variability.}
\label{fig:vis2}
\vspace{-2pt}
\end{figure}

\begin{table}[t]
\centering
\captionsetup[table]{aboveskip=2pt, skip=3pt}
\setlength{\tabcolsep}{4pt}
\renewcommand{\arraystretch}{0.92}
\small
\caption{Ablation on router types (Pearson $r$; $S_{1}$--$S_{5}$ mean).}
\label{tab:router_type_ablation}
\begin{tabular}{c S[table-format=1.3] S[table-format=1.3] S[table-format=1.3]}
\toprule
\textbf{Router Types} & \textbf{TRIBE} & \textbf{ImageBind} & \textbf{Qwen2.5-Omni} \\
\midrule
Only Token Router  & 0.176 & 0.131 & 0.107 \\
Only Prior Router  & 0.248 & 0.205 & 0.173 \\
\rowcolor{gray!15}
Both               & \textbf{0.273} & \textbf{0.221} & \textbf{0.220} \\
\bottomrule
\end{tabular}
\normalsize
\end{table}

Tab.~\ref{tab:improvements} shows that MIND outperforms both an MLP decoder (single-path projection) and an input-driven MMoE (Multi-Gate Mixture-of-Experts with learned token gating~\cite{ma2018modeling}) across heterogeneous fusion backbones.
This supports the plug-and-improve claim: gains persist despite differences, indicating that MIND improves cross-subject generalization rather than overfitting to a specific fusion design.
Mechanistically, AFIRE provides a standardized post-fusion token interface, while the subject-prior router calibrates subject-level gain; Top-$K$ sparse routing with load-balancing regularizes expert usage, yielding a more robust trade-off than token-only gating in MMoE under backbone shifts.

\vspace{-0.5em}
\subsection{Ablation Studies}
\vspace{-0.3em}

\noindent\textbf{Router Components.}
Tab.~\ref{tab:router_type_ablation} compares a token-only router, a prior-only router, and their combination. Neither component alone suffices; the combined router achieves the best results across backbones, suggesting that transient (token-dependent) and persistent (subject-prior) factors are complementary and should be integrated via sparse routing.

\vspace{-0.7em}
\subsection{Visualizations and Analysis}
\vspace{-0.3em}
\noindent\textbf{Parcel-wise Correlation across Backbones.}
Fig.~\ref{fig:vis1} shows parcel-wise Pearson $r$ maps for TRIBE, ImageBind, and Qwen2.5-Omni (all decoded with \textit{MIND}); similar spatial patterns and overlapping high-$r$ regions across backbones support AFIRE’s fusion-agnostic design and \textit{MIND}’s plug-and-play behavior.

\noindent\textbf{Subject Routing Dynamics.}
Fig.~\ref{fig:vis2} unfolds time-resolved routing weights over the first 100 TRs for $S_{1}$/$S_{2}$/$S_{3}$/$S_{5}$ on the same episode, showing subject-aware expert preferences from the prior combined with token-dependent gating, consistent with the ISG gains in Tab.~\ref{tab:improvements}.

\vspace{-1em}
\section{Conclusion}
\vspace{-0.5em}
We introduced \textit{AFIRE}, an agnostic post-fusion interface, and \textit{MIND}, a plug-and-play Mixture-of-Experts decoder with a subject-aware gate, for end-to-end whole-brain prediction. The framework consistently outperforms strong baselines across diverse backbones and subjects, improving cross-subject generalization while yielding interpretable expert allocation under backbone shifts. Remaining limitations include reliance on post-fusion token quality and added MoE computation. Future work will jointly optimize encoders with AFIRE, enhance robustness to missing modalities and variable TRs, and pursue privacy-preserving personalization and uncertainty-aware encoding.

\bibliographystyle{IEEEbib}
\bibliography{strings,refs}

\end{document}